\documentclass[twoside,11pt]{article}

\usepackage{jmlr2e}

\usepackage{booktabs} 

\usepackage{amsmath}
\usepackage{amssymb}
\usepackage{mathtools}
\usepackage{amsthm}
\usepackage{bbm}
\usepackage{algorithm}
\usepackage{algorithmic}
\usepackage{siunitx}

\usepackage{makecell}  
\usepackage{multirow}  
\usepackage{graphicx}  
\usepackage{xcolor}
\usepackage{subcaption}

\DeclareMathOperator*\argmax{arg\,max}

\jmlrheading{1}{}{}{}{}{}

\firstpageno{1}

\begin{document}

\title{Practical Path-based Bayesian Optimization}

\author{\name Jose Pablo Folch \email jose.folch16@imperial.ac.uk \\
       \addr Department of Mathematics, Imperial College London
       \AND
       \name James Odgers \email james.odgers16@imperial.ac.uk \\
       \addr Department of Computing, Imperial College London
       \AND
       \name Shiqiang Zhang \email s.zhang21@imperial.ac.uk \\
       \addr Department of Computing, Imperial College London
       \AND
       \name Robert M Lee \email robert-matthew.lee@basf.com \\
       \addr BASF SE
       \AND
       \name Behrang Shafei \email behrang.shafei@basf.com \\
       \addr BASF SE
       \AND
       \name David Walz \email david-simon.walz@basf.com \\
       \addr BASF SE
       \AND
       \name Calvin Tsay \email c.tsay@imperial.ac.uk \\
       \addr Department of Computing, Imperial College London
       \AND
       \name Mark van der Wilk \email m.vdwilk@imperial.ac.uk \\
       \addr Department of Computing, Imperial College London
       \AND
       \name Ruth Misener \email r.misener@imperial.ac.uk \\
       \addr Department of Computing, Imperial College London
       }


\maketitle

\begin{abstract}
There has been a surge in interest in data-driven experimental design with applications to chemical engineering and drug manufacturing. Bayesian optimization (BO) has proven to be adaptable to such cases, since we can model the reactions of interest as expensive black-box functions. Sometimes, the cost of this black-box functions can be separated into two parts: (a) the cost of the experiment itself, and (b) the cost of changing the input parameters. In this short paper, we extend the SnAKe algorithm to deal with both types of costs simultaneously. We further propose extensions to the case of a maximum allowable input change, as well as to the multi-objective setting.
\end{abstract}


\section{Introduction and related work} \label{sec: intro}

Efficiently optimizing the desired qualities of materials or chemicals in experimental settings is important, given the costliness of experiments. Bayesian Optimization (BO) \citep{jones1998efficient, shahriari2016bo}  has emerged as a potent tool for this purpose, e.g., see \cite{griffiths2020constrained}. However, the diversity of experimental setups necessitates extending current methods to accommodate common constraints and objectives.

BO seeks to optimize an expensive black-box function through the use of probabilistic models, usually Gaussian Processes (GPs) \citep{rasmussen2005gps}, which are then used to select inputs in sequential fashion. This usually involves optimizing a tractable acquisition function, $\alpha(x)$, so that the selected input is given by:
\begin{equation}
	x_{t+1} = \argmax_{x \in \mathcal{X}} \alpha(x | D_t)
\end{equation}
where $D_t$ is the data-set at iteration $t$, and $x_i$ is the $i$-th experiment.

Recent work has focused on path-based BO, where the cost of changing inputs between subsequent experiments must be taken into account. This surge has been inspired by scientific applications to environmental monitoring \citep{bogunovic2016truncated, mutny2023active, folch2022snake, yang2023mongoose, samaniego2021bayesian}, chemistry \citep{folch2022snake, vellanki2017proccess-constrained}, and energy systems \citep{ramesh2022movement}.

Approaches for path-based BO have been considered through meta-learning \citep{yang2023mongoose}, and in the contextual BO setting \citep{ramesh2022movement}. \citet{mutny2023active} approach a similar problem by considering the optimization of trajectories in active learning. Our recent work \citep{folch2022snake} proposes SnAKe, a path-based variation of BO that is able to minimize the cumulative cost of input changes, by iteratively: (i) creating large batches of possible experiments using Thompson Sampling \citep{pmlr-v84-kandasamy18a}, (ii) solving a Travelling Salesman Problem to order the experiments in a way that minimizes input costs, and (iii) updating the path as observations arrive, using a `point-deletion' heuristic that helps avoid getting stuck in local optima.

SnAKe is limited in some practical aspects. This work presents three extensions:
\begin{itemize}
	\item[I.] SnAKe requires the total number of experiments to be selected \textit{a priori}, but, in practice we might not know how many experiments are needed, or we might want to minimize the number of experiments, e.g., to save on materials. 
	\item[II.] Even though SnAKe minimizes cumulative input changes, individual input changes between iterations are unbounded. This could pose a practical problem in some applications, where large changes may be unsafe or infeasible. For example, there may be limitations on maximum temperature or the rate of temperature change.  
	\item[III.] The algorithm assumes a single objective function. In many applications, experimentalists are interested in problems with many objectives. In multi-objective settings, we seek to find a Pareto front of solutions as opposed to optimizing a single function.
\end{itemize}

Other limitations of SnAKe include poor performance in high-dimensional settings, or when the experimental budget is low ($< 100$). \citet{yang2023mongoose} introduce MONGOOSE, addressing both problems using meta-learning. 

Further related work includes \citet{nguyen2017regret}, who explore using the Expected Improvement (EI) acquisition function as a stopping criteria, similar to the use of Expected Improvement per unit cost (EIpu)~\citep{snoek2012practical} which we introduce here. \citet{paulson2022localconstraints} and \citet{paulson2023lsr} enable incorporation of local movement constraints in BO using approximate dynamic programming, and by combining local and global steps depending on the local expected improvement. \citet{samaniego2021bayesian} explore the use of truncation for minimizing movement costs under the context of environmental monitoring. Multi-objective optimization has seen recent research interest, in particular at the intersection of chemistry and machine learning \citep{thebelt2022multi, tu2022joint}.

\section{Practical Path-based Bayesian optimization}

\subsection{Self-stopping SnAKe}

\begin{algorithm}
\caption{Expected Improvement Point Deletion} \label{algo: EI_Point_Deletion}
\begin{algorithmic}
 \STATE INPUT: Proposed batch, $\mathcal{P
 }_t$, of size $(T - t)$. Model-misspecification safety parameter $\nu$, termination parameter $\delta$.
 \FOR{$x_p \in \mathcal{P}_t$}
 \IF{( $EI(x_p) / C^{(0)}(x_p) < \delta$ ) and ( Var$[\ x_p \ | \ D_t \ ]  < \nu$ )}
 \STATE $\mathcal{P}_t \leftarrow \mathcal{P}_t \backslash \{x_p\}$
 \ENDIF
\ENDFOR
\IF{$|\mathcal{P}_t| = 0$}
\STATE Terminate optimization
\ENDIF
\STATE OUTPUT: Reduced Batch, $\tilde{\mathcal{P}}$, of unknown size. Optimization termination criteria.
\end{algorithmic}
\end{algorithm}

We begin by investigating the first problem. We assume that we want to optimize an expensive black-box function $f$ and that the cost of experiment $t+1$ can be described by:
\begin{equation}
	C(x_t, x_{t+1}) = C^{(0)}(x_{t+1}) + C^{(\Delta)}(x_t, x_{t+1})
\end{equation}
where $x_{t+1}$ is the next experimental input we want to query, $x_t$ is our current input, $C^{\Delta}(x_{t}, x_{t+1})$ is an input-change cost function equal to zero when $x_{t+1} = x_t$, and $C^{(0)}(x_{t+1})$ represents the fixed cost of each experiment.
The goal is to optimize the function for the smallest possible total cost. Finding an exact optimum of a black-box function is practically impossible; therefore the BO loop should terminate once we have a good approximation.

To combine BO stopping conditions and SnAKe, we modify the Point Deletion criteria~\cite[Section~3.7]{folch2022snake} using EIpu \citep{snoek2012practical} as a stopping condition. For a certain point $x$, if the EIpu, $EI(x) / C^{(0)}(x)$, is less than a given threshold $\delta>0$, then we delete it.
We also add an extra criterion based on the prediction variance of the point to avoid early termination in the case where the model is misspecified: we will not delete points that still have significant uncertainty. Using this same criteria (to avoid model misspecification), we were able to significantly boost EIpu's performance, achieving comparable results to SnAKe. We detail the method in Algorithm \ref{algo: EI_Point_Deletion}.

\subsection{Truncated SnAKe}

We now investigate the problem of maximum allowable input changes, motivated by, e.g., safety or physical limits. 
In particular, we assume the optimization now involves the \textit{rate-of-change} constraint: 
\begin{equation}
	| x_{t+1} - x_t | < \delta_\mathit{MAX}, \quad \forall \ t \in \{1,..., T\}
\end{equation}
We modify the criteria that SnAKe uses when selecting which experiment to query next: truncate SnAKe's input change whenever it exceeds $\delta_\mathit{MAX}$. The modification is shown in Algorithm \ref{algo: TrSnAKe}, where $[v]^{(u)}$ denotes the unit-length vector in direction $v$.

\begin{algorithm}
\caption{Truncated SnAKe (TrSnAKe)} \label{algo: TrSnAKe}
\begin{algorithmic}
 \STATE Run SnAKe as normal with the following extra criterion for each new experiment, $x_{t+1}$:
 \IF{$|x_{t+1} - x_t| > \delta_\mathit{MAX}$}
  \STATE $x_{t+1} \leftarrow x_t + [x_{t+1} - x_t]^{(u)}\delta_\mathit{MAX}$
 \ENDIF
\end{algorithmic}
\end{algorithm}

\subsection{Multi-objective SnAKe}

We consider a simple multi-objective extension of SnAKe by using random scalarizations of the objective functions \citep{paria2020flexible}. We define the problem as trying to find the Pareto front of a set of objective functions $\{ f^{(k)}(x) \}_{k=1}^K$. The Pareto front (in the maximization setting) is defined as the set of points where we cannot increase the value of an objective without decreasing the value of another.

SnAKe creates a large batch of points using Thompson Sampling, which is then considered for evaluation and ordered in such a way that input costs are minimized. \citet{paria2020flexible} give a multi-objective extension of Thompson Sampling, where we consider a random parametric scalarization of the objectives as a single objective function.

Assume we have $K$ objectives, each modelled by an independent GP. We choose a scalarization function $\mathcal{S}(\{f^{(k)}\}_{k=1}^K, \lambda)$ where $\lambda$ is a random parameter with distribution $p(\lambda)$. Specifically, we test a linear scalarization function $\mathcal{S}(\{f^{(k)}\}_{k=1}^K, \lambda)(x) = \sum_{k=1}^K \lambda_k f^{(k)}(x)$ and Tchebyshev scalarization $\mathcal{S}(\{f^{(k)}\}_{k=1}^K, \lambda)(x) = \min_{k = 1, ..., K} \lambda_k (f^{(k)}(x) - z^{(k)})$, where $z^{(k)}$ are some reference points \citep{nakayama2009sequential}. The latter can handle non-convex Pareto fronts. To create the batch, we simply sample both the independent GPs and $\lambda \sim p(\lambda)$ and then optimize the single objective (one for each GP sample). We choose this particular multi-objective method for its computational simplicity, which is required when creating large batches. We outline the procedure in Algorithm \ref{algo: MOSnAKe}.

\begin{algorithm}
\caption{Multi-objective SnAKe (MO-SnAKe)} \label{algo: MOSnAKe}
\begin{algorithmic}
\STATE Run SnAKe as normal, but replace the Thompson Sampling in the batch creation step with Random Scalarization Thompson Sampling. That is, create the batch with the following algorithm: \\
\FOR{$i = 1, ..., T$}
\STATE $f^{(1)}, ..., f^{(K)} \sim GP(\mu_t, \sigma_t | D_t)$ \\
\STATE $\lambda \sim p(\lambda)$ \\
\STATE $x_{t+1}^{(i)} = \argmax_{x \in \mathcal{X}}{\mathcal{S}(\{f^{(k)}\}_{k=1}^K, \lambda})(x)$ \\
\ENDFOR
\end{algorithmic}
\end{algorithm}

\section{Experimental Results}

All implementations use GPyTorch \citep{gardner2018gpytorch} and BoTorch \citep{balandat2020botorch}. To implement SnAKe, we use the non-parametric version of point-deletion when applicable (denoted by an $\ell$ prefix). In all plots, we zoom into the area of interest (low-cost region) and plot $\pm$0.5 the standard deviations of 25 runs with random initial conditions. Dashed vertical lines in plots show the average cost used. As in \cite{folch2022snake}, the hyper-parameters of the GP are optimized by maximizing the marginal-likelihood every 25 iterations. Random refers to uniform samples ordered by solving the Travelling Salesman Problem. We highlight a few benchmarks here and include further studies in the Appendix.

We tested self-stopping SnAKe (ssSnAKe) in three synthetic benchmarks: Branin2D, Hartmann3D and Hartmann6D. We compare against SnAKe, vanilla EIpu, and a new implementation of EIpu that also considers model certainty as termination criteria.

\begin{figure}[!hb]
    \centering
    \includegraphics[width = 0.9 \textwidth]{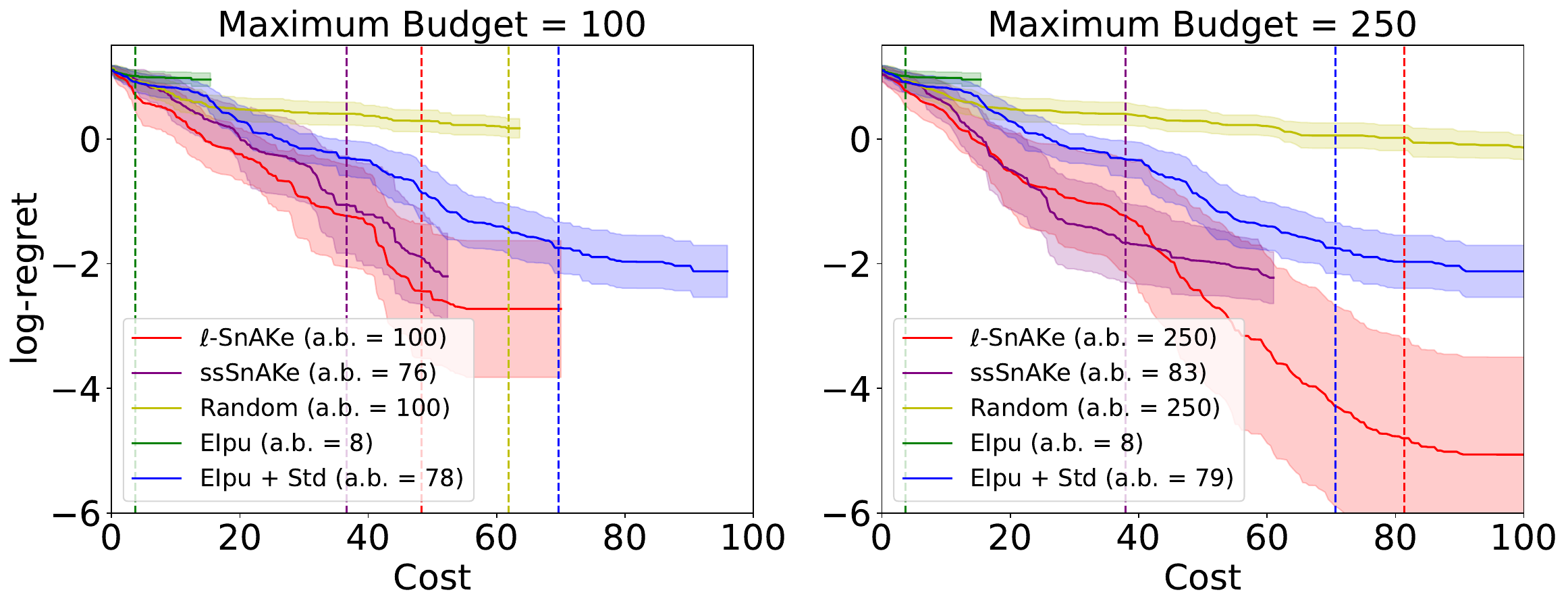}
    \vspace{-2mm}
    \caption{Self-stopping optimization of Hartmann 6D with different maximum budgets. Vanilla EIpu stops too early due to model misspecification; our modification that includes uncertainty criteria greatly improves performance. The average budget used (a.b.) is robust to maximum budget choice.} 
    \label{fig: main_paper_ss}
\end{figure}

SnAKe achieves the best regret; however, it expends the entire budget and therefore incurs high costs for very small gains. ssSnAKe achieves similar performance to SnAKe up to the point where it automatically terminates. Our modification to EIpu significantly improves the performance, achieving the best results in Hartmann3D and comparable results in Branin2D benchmarks. It is good to note that the ssSnAKe seems robust to the maximum budget, using approximately the same number of experiments, even when the maximum budget is increased. We showcase the Hartmann6D benchmark in Figure \ref{fig: main_paper_ss}.

For rate-of-change constraint problems, we test on two synthetic benchmarks: Branin2D and Hartmann3D. We compare Truncated SnAKe (TrSnAKe), Truncated Expected Improvement (TrEI) \citep{samaniego2021bayesian}, SnAKe, and EIpu. We emulate movement constraints using a cost function with a jump when the movement is larger than $\delta_{MAX}$:
\begin{equation*}
    \mathcal{C}(x_t, x_{t+1}) = 0.2 || x_{t+1} - x_t ||_2 + \mathbbm{1}(|| x_{t+1} - x_t ||_2 > \delta_{MAX})
\end{equation*}

\begin{figure}[ht]
    \centering
    \includegraphics[width = 0.9\textwidth]{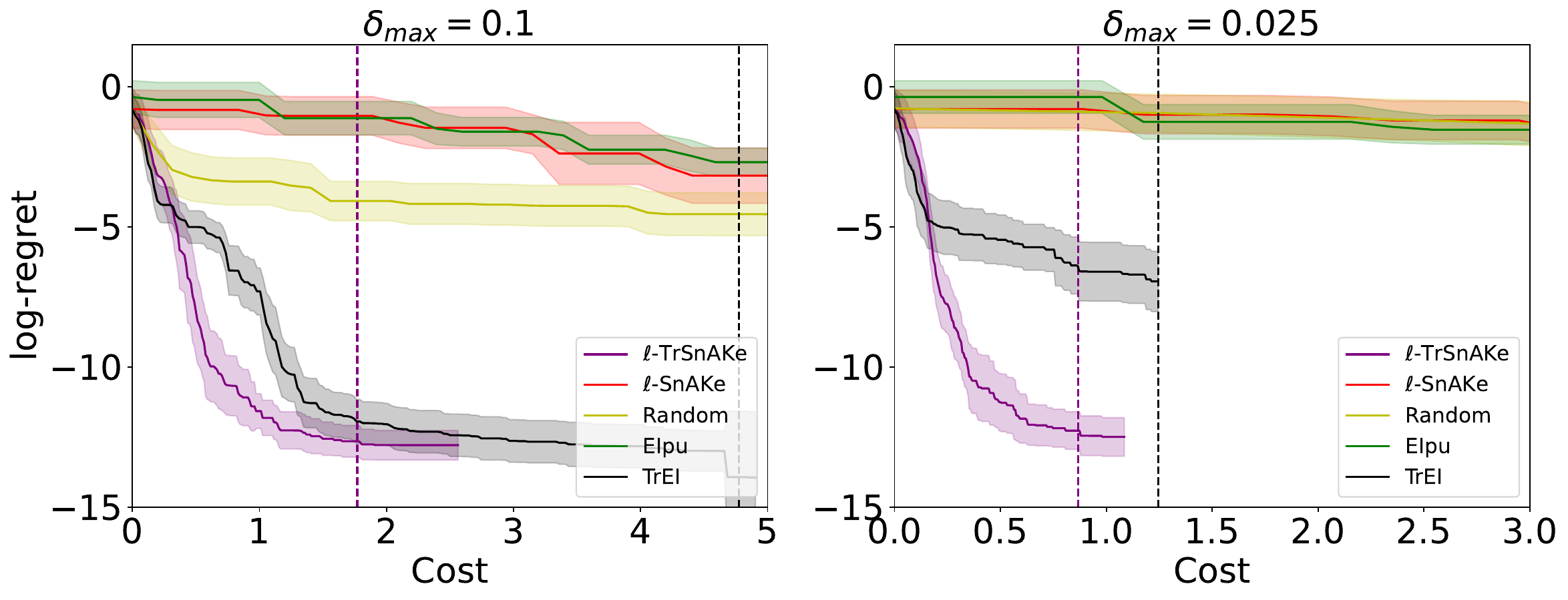}
    \vspace{-2mm}
    \caption{Results for truncated Branin2D with a budget of 250 queries. The cost of non-truncated methods is very high due to penalization for breaking movement constraints. TrSnAKe achieves the best performance for low cost.}
    \label{fig: main_paper_truncated}
\end{figure}

For each benchmark, we test with $\delta_{MAX} \in \{ 0.1, 0.025 \}$. TrSnAKe performs best in three out of four cases. Methods that cannot consider movement constraints are heavily penalized and achieve poor performance. Figure \ref{fig: main_paper_truncated} shows the results for the Branin2D benchmark.

Finally, we test multi-objective SnAKe (MO-SnAKe) on two benchmarks: the Schekel function, which was used to model pollution in Ypacarai Lake \citep{samaniego2021bayesian}, and the SnAr chemistry benchmark, implemented in Summit \citep{felton2021summit}. We compare MO-SnAKe and Scalarized Thompson Sampling \citep{paria2020flexible}. We note that \citet{folch2022snake} used SnAr as a real-world application; however, they restricted the problem to single objective. This work is able to consider the original multi-objective problem. 

\begin{table*}[htp]
    \caption{Results of multi-objective optimization experiments on four dimensional benchmarks. \textcolor{red}{Red} and \textcolor{blue}{blue} indicate linear and Tchebyshev scalarazations respectively.}
    \centering
    \vspace{-2mm}
    \begin{tabular}{ccr@{${}\pm{}$}rr@{${}\pm{}$}rr@{${}\pm{}$}rr@{${}\pm{}$}r}
        \toprule
        & Method & \multicolumn{2}{c}{GD} & \multicolumn{2}{c}{IGD} & \multicolumn{2}{c}{MPFE} & \multicolumn{2}{c}{Cost} \\
        \toprule
        \multirow{5}{*}{\rotatebox[origin=c]{90}{Schekel}} 
        & \textcolor{red}{$\ell$-MO-SnAKe}    & \textbf{0.05} & 0.06 & \textbf{0.09} & 0.07 & \textbf{0.23} & 0.08 & \textbf{9.60} & 6.12 \\
        & \textcolor{blue}{$\ell$-MO-SnAKe} & 0.07 & 0.05 & 0.15 & 0.05 & 0.29 & 0.07 & 11.24 & 9.94 \\
        & \textcolor{red}{TS}                 & \textbf{0.05} & 0.04 & 0.11 & 0.07 & 0.24 & 0.08 & 57.21 & 12.97 \\
        & \textcolor{blue}{TS}              & 0.06 & 0.04 & 0.16 & 0.05 & 0.30 & 0.07 & 63.18 & 12.30 \\
        & Random                      & 0.22 & 0.05 & 0.22 & 0.04 & 0.31 & 0.05 & 59.62 & 1.08\\
        \midrule
        \multirow{5}{*}{\rotatebox[origin=c]{90}{SnAr}} 
        & \textcolor{red}{$\ell$-MO-SnAKe}    & 0.00040 & 0.00015 & 0.19 & 0.06 & 0.67 & 0.22 & 614.94 & 229.98 \\
        & \textcolor{blue}{$\ell$-MO-SnAKe} & 0.00044 & 0.00020 & 0.18 & 0.07 & 0.65 & 0.26 & \textbf{478.43} & 219.16 \\
        & \textcolor{red}{TS}                 & 0.00035 & 0.00008 & 0.18 & 0.06 & 0.64 & 0.22 & 1967.56 & 447.31 \\
        & \textcolor{blue}{TS}              & \textbf{0.00031} & 0.00008 & 0.18 & 0.07 & 0.63 & 0.25 & 1602.68 & 401.3 \\
        & Random                      & 0.02498 & 0.01157 & \textbf{0.16} & 0.04 & \textbf{0.35} & 0.15 & 1195.68 & 42.07 \\
        \bottomrule 
    \end{tabular}
    \label{tab: mo_results}
\end{table*}

For performance metrics, we use Euclidean generational distance (GD) which measures the average distance from the the approximated Pareto front to the true one, inverted generational distance (IGD) which measure the average distance from the true Pareto front to the approximated one, and the maximum Pareto frontier error (MPFE) which measures the largest error from the approximated Pareto frontier to the true one. We obtain a `true' set of Pareto points by running \texttt{NSGA-II} \citep{deb2002fast}. Table \ref{tab: mo_results} shows MO-SnAKe performs comparable to Scalarized Thompson Sampling \citep{paria2020flexible} at much lower cost.

\section{Conclusion}

In this paper, we explore extensions of SnAKe to path-based BO, to make it feasible to more practical situations. For each situation, we mention areas where the particular modification might be useful, and empirically show the advantages of using our modifications.

\vskip 0.2in
\bibliography{sample}

\newpage
\appendix
\onecolumn
\section{Acknowledgements}

 JPF is funded by EPSRC through the Modern Statistics and Statistical Machine Learning (StatML) CDT (grant no. EP/S023151/1) and by BASF SE, Ludwigshafen am Rhein. SZ was supported by an Imperial College Hans Rausing PhD Scholarship.  We would also like to thank Daniel Lengyel and his ChatGPT subscription for helping us format the tables.

\section{Complete numerical results}

In this section we include further experimental results and more implementation details. As per \cite{folch2022snake}, we implement Thompson Sampling using the efficient sampling procedure introduced by \citet{wilson2020efficiently}, solve the Travelling Salesman using Simulated Annealing \citep{kirkpatrick1983simmulated} and the NetworkX package \citep{hagberg2008networkx}. Optimizations of the Thompson Samples are done in PyTorch \citep{paszke2019pytorch} using Adam \citep{kingma2014adam}. The code used to produce the results will be available after peer-review in the original SnAKe repository (\url{https://github.com/cog-imperial/SnAKe}).

\subsection{Early stopping experiments}

\begin{figure}[ht]
    \centering
    
    \begin{subfigure}[t]{0.85\textwidth}
	\includegraphics[width = \textwidth]{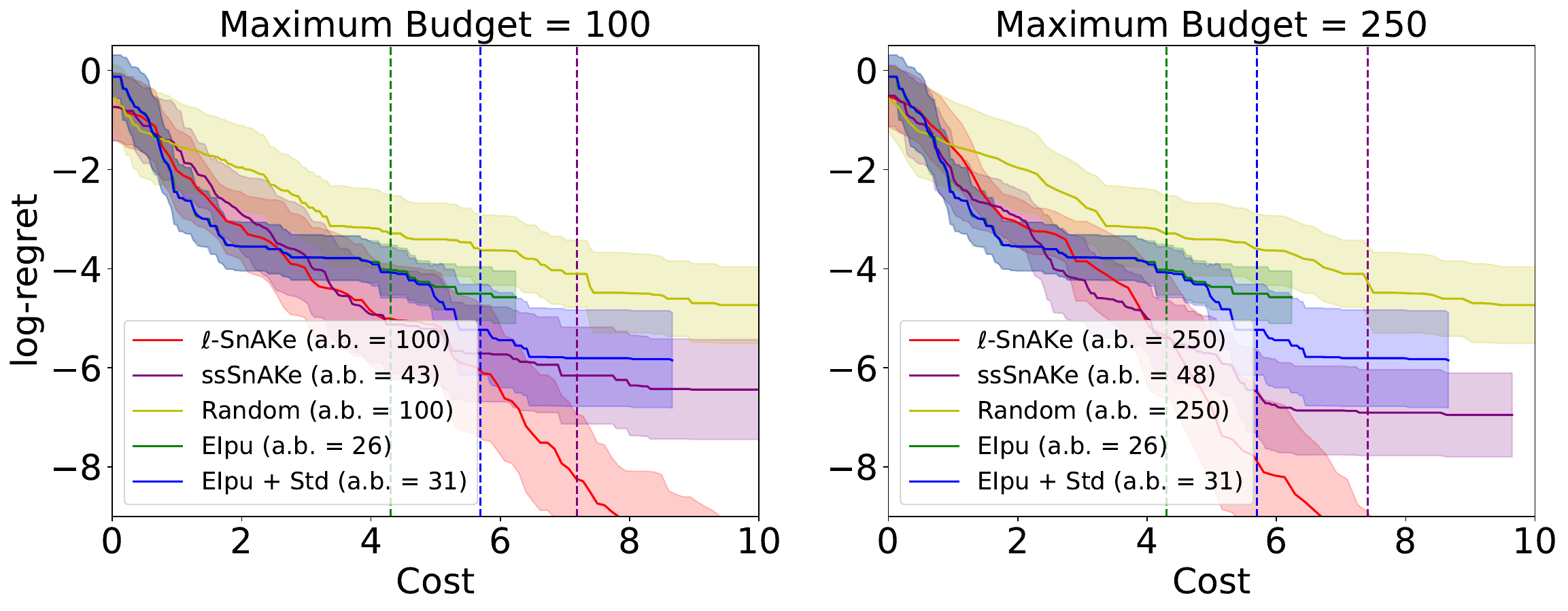}
	\caption{Branin2D}
	\label{fig: ss_branin2d} 
    \hspace{10mm}
	\end{subfigure}
    \begin{subfigure}[t]{0.85\textwidth}
	\includegraphics[width = \textwidth]{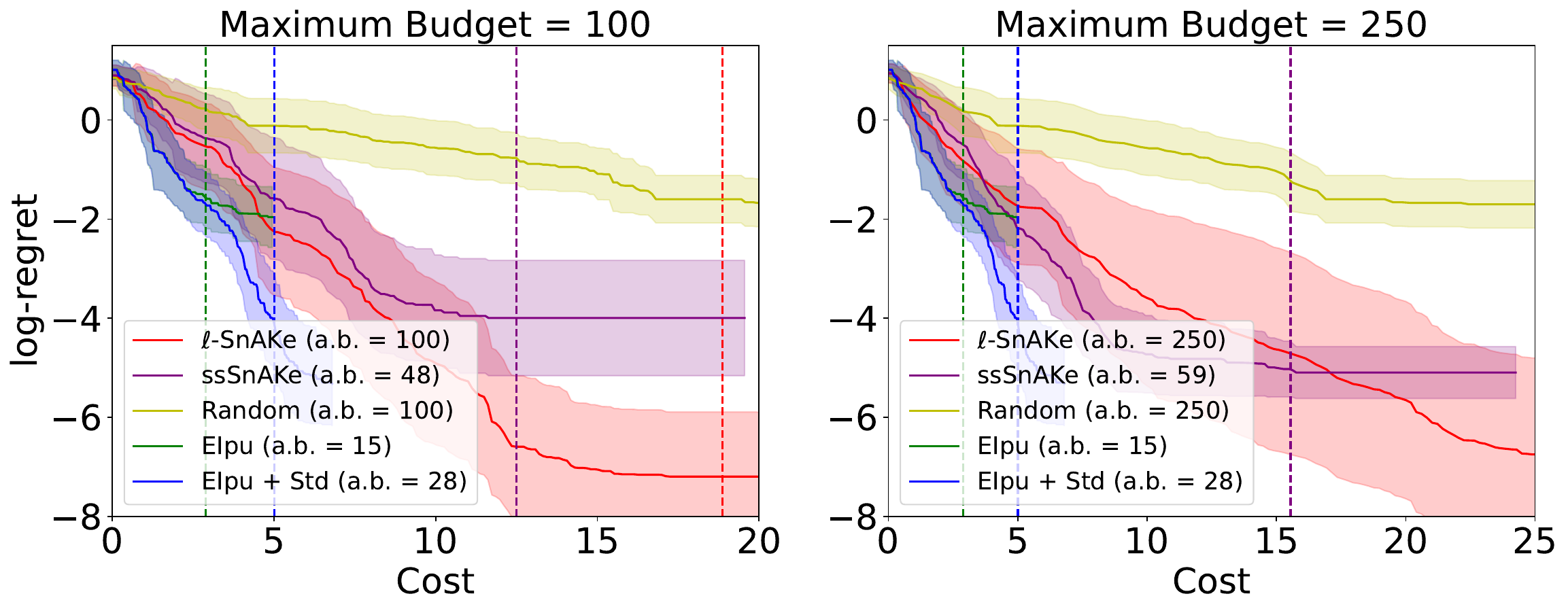}
	\caption{Hartmann3D}
	\label{fig: ss_hartmann3d}
	\end{subfigure}
    \caption{Results of self-stopping experiment on remaining synthetic benchmarks. We show average budget (a.b.) used in the captions, and the average cost as vertical lines. In all cases vanilla EIpu seems to terminate too early. Once we have included the uncertainty reduction criteria (EIpu+Std), the performance improves significantly. EIpu+Std achieves the best performance in Hartmann3D, and comparable performance to self-stopping SnAKe (ssSnAKe) in Branin2D. In all cases ssSnAKe follows the performance of SnAKe closely, up to the point of termination. In all cases ssSnAKe and EIpu+Std appear to be robust to the maximum budget, terminating the optimizaiton at approximately the same time even when the maximum budget is increased.}
    \label{fig: appendix_ss_results}
\end{figure}

We tested self-terminating SnAKe on three total benchmarks. Hartmann 6D can be seen in Figure \ref{fig: main_paper_ss}. The cost function used in the problem is given by:
\begin{equation}
    \mathcal{C}(x_t, x_{t+1}) = \alpha + || x_{t+1} - x_t ||_2
\end{equation}
Where $\alpha = 0.05, 0.1, 0.25$ in each of Branin2D, Hartmann3D, and Hartmann6D respectively. Figure \ref{fig: appendix_ss_results} shows the results of the remaining two benchmarks. ssSnAKe gives the strongest performance in Hartmann6D, EIpu with the additional variance stopping condition is strongest in Hartmann3D, and the methods are comparable in Branin2D.

\subsection{Truncated Experiments}

We further tested on Hartmann3D for the experiments with rate-of-change constraints. In this case Truncated SnAKe (TrSnAKe) achieves stronger results when the rate of change is larger, giving it more freedom to explore the space. Truncated Expected Improvement (TrEI) gives better results when restricted to a smaller constraint.

\begin{figure}[t]
    \centering
    \includegraphics[width = 0.85\textwidth]{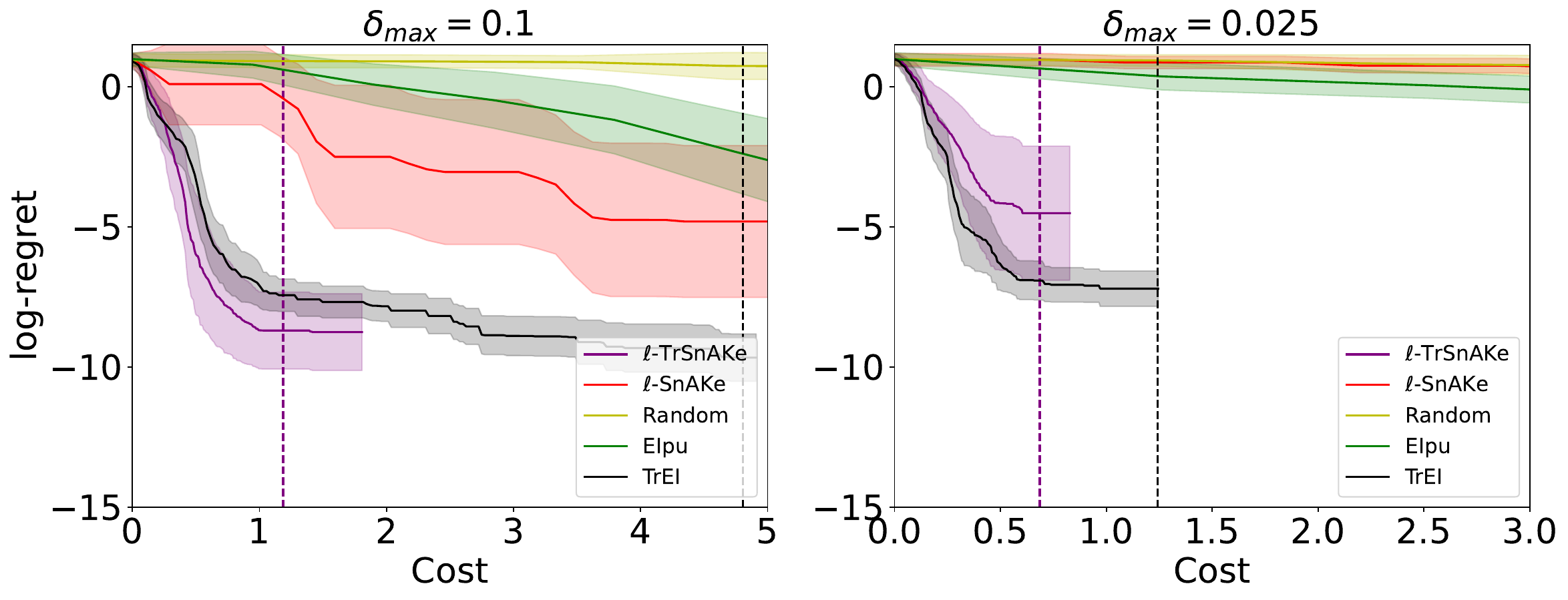}
    \caption{Results for truncated Hartmann3D. The cost of non-truncated methods is very high due to constantly being penalized for breaking movement constraints. We show results for a budget of 250 queries, with two different constraint sizes. TrSnAKe achieves the best performance when the movement constraint is larger, and TrEI the best when the movement constraint is lower.}
    \label{fig: appendix_truncated_results}
\end{figure}

\begin{table*}[h!]
    \caption{Results of multi-objective optimization experiments on four dimensional benchmarks with relaxed Pareto front. Even though the approximated Pareto front now contains observations known to \textit{not} be Pareto optimal, the performance in the IGD and MPFE metrics improves. \textcolor{red}{Red} and \textcolor{blue}{blue} indicate linear and Tchebyshev scalarazations respectively.}
    \centering
    \vspace{1mm}
    \begin{tabular}{ccr@{${}\pm{}$}rr@{${}\pm{}$}rr@{${}\pm{}$}rr@{${}\pm{}$}r}
    \toprule
    Problem & Method & \multicolumn{2}{c}{GD} & \multicolumn{2}{c}{IGD} & \multicolumn{2}{c}{MPFE} & \multicolumn{2}{c}{Cost} \\
    \toprule
            & \textcolor{red}{$\ell$-MO-SnAKe}    & \textbf{0.05} & 0.05 & \textbf{0.09} & 0.07 & \textbf{0.22} & 0.09 & \textbf{9.60} & 6.12 \\
            & \textcolor{blue}{$\ell$-MO-SnAKe} & 0.06 & 0.04 & 0.15 & 0.05 & 0.28 & 0.07 & 11.24 & 9.94 \\
    Schekel & \textcolor{red}{TS}         & \textbf{0.05} & 0.04 & 0.11 & 0.07 & 0.24 & 0.08 & 57.21 & 12.97 \\
            & \textcolor{blue}{TS}      & 0.06 & 0.04 & 0.16 & 0.05 & 0.30 & 0.08 & 63.18 & 12.30 \\
            & Random              & 0.22 & 0.05 & 0.22 & 0.04 & 0.31 & 0.05 & 59.62 & 1.08\\
    \midrule
            & \textcolor{red}{$\ell$-MO-SnAKe}    & 0.00248 & 0.00118 & \textbf{0.06} & 0.02 & \textbf{0.24} & 0.08 & 614.94 & 229.98 \\
            & \textcolor{blue}{$\ell$-MO-SnAKe} & 0.00262 & 0.00106 & \textbf{0.06} & 0.04 & 0.27 & 0.15 & \textbf{478.43} & 219.16 \\
    SnAr    & \textcolor{red}{TS}           & 0.00214 & 0.00091 & 0.07 & 0.03 & 0.29 & 0.08 & 1967.56 & 447.31 \\
            & \textcolor{blue}{TS}        & \textbf{0.00171} & 0.00047 & \textbf{0.06} & 0.02 & \textbf{0.24} & 0.08 & 1602.68 & 401.3 \\
            & Random              & 0.02725 & 0.01345 & 0.14 & 0.04 & 0.25 & 0.09 & 1195.68 & 42.07 \\
    \bottomrule 
    \end{tabular}
    \label{tab: mo_results_appendix}
\end{table*}

\subsection{Multi-objective experiments}

Finally, we consider a relaxed Pareto front to explain why Random appears to outperform BO in IGD and MPFE in the SnAr chemistry benchmark. This is due to the Pareto front being almost a horizontal line, and therefore the algorithms prefer to explore the part of the line that maximizes the second objective. We consider a relaxed Pareto front by taking into account all points where we cannot improve on our current objective without decreasing the second objective by more than 0.01 (before this limit was strictly zero).

Using this definition for our approximated Pareto front, Table \ref{tab: mo_results_appendix} shows improved performance in IGD and MPFE, at the expense of a decrese in performance in GD (as now point in the approximated Pareto front may not be Pareto optimal) showcasing the difficulty in measuring good performance in multi-objective optimization.


\end{document}